\documentclass[accepted]{article}
\usepackage{diagbox}
\usepackage{microtype}
\usepackage{graphicx}
\usepackage{subfigure}
\usepackage{booktabs}
\usepackage{hyperref}
\usepackage{icml2024}
\usepackage{enumitem}
\usepackage{amsmath}
\usepackage{amssymb}
\usepackage{mathtools}
\usepackage{amsthm}
\usepackage[capitalize,noabbrev]{cleveref}
\theoremstyle{plain}

\theoremstyle{definition}

\theoremstyle{remark}

\usepackage[textsize=tiny]{todonotes}
\usepackage[T1]{fontenc}
\icmltitlerunning{Submission and Formatting Instructions for ICML 2024}
\usepackage[font=small,skip=4pt]{caption}
\usepackage{listings}
\begin{document}
\twocolumn[

\icmltitle{LoMA: Lossless Compressed Memory Attention}
\icmlsetsymbol{equal}{*}

\begin{icmlauthorlist}
\icmlauthor{Yumeng Wang\textsuperscript{*}}{dg}
\icmlauthor{Zhenyang Xiao\textsuperscript{*}}{dg,pku}
\end{icmlauthorlist}

\icmlaffiliation{dg}{Deepglint Inc.}
\icmlaffiliation{pku}{Peking University}

\icmlcorrespondingauthor{Yumeng Wang}{yumengwang@deepglint.com}

\icmlkeywords{Artificial Intelligence, Language Model}

\vskip 0.3in
]

\printAffiliationsAndNotice{\icmlEqualContribution}

\begin{abstract}
Large Language Models (LLMs) face limitations due to the high demand on GPU memory and computational resources when handling long contexts. While sparsify the Key-Value (KV) cache of transformer model is a typical strategy to alleviate resource usage, it unavoidably results in the loss of information. We introduce Lossless Compressed Memory Attention (LoMA), a novel approach that enables lossless compression of the KV cache, thereby reducing the memory and computational demands during autoregressive generation. LoMA incorporates a specialized training or fine-tuning precedure alongside an autoregressive generation algorithm optimized for the compressed context. Our method compresses the KV cache after every $tc$ generated tokens with a compression ratio of $c$ and a target compressed length $t$, and this process occurs within a single inference pass without dependency on auxiliary models. We engineered an efficient training scheme involving specific inputs, attention masks, and position identifiers to instill this compression capability. Experimental validation has demonstrated that LoMA significantly reducing computational consumption and memory usage through achieving lossless KV cache compression.
\end{abstract}

\begin{figure*}[]
    \centering
    \includegraphics[width=\textwidth]{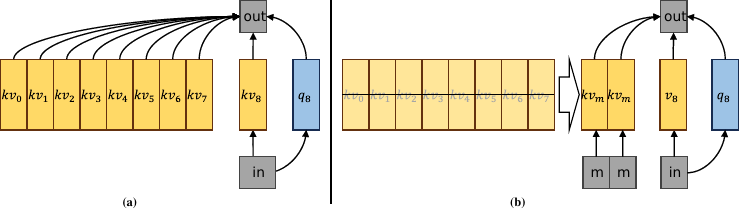}
    \caption{Comparison of the standard transformer model with the LoMA model in autoregressive generation: (a) In the standard transformer model's autoregressive generation, the input token and the previous context's KV cache are fed together into the attention module to compute and predict the next token. (b) In the LoMA model's autoregressive generation, the previous context's KV cache is first compressed, and the input token is processed with the compressed KV cache by the attention module.}
    \label{fig_6}
\end{figure*}

\section{Introduction}

In the field of Natural Language Processing (NLP), understanding and managing long context represents one of the significant challenges for achieving in-depth language comprehension. Research into long context not only enhances the model's capabilities in processing lengthy dialogues, document comprehension, and information retrieval tasks but also aids in achieving more precise language inference and knowledge extraction, thereby facilitating progress in applications such as machine translation, summarization, and question-answering systems\cite{yang_baichuan_2023}. In these tasks, users expect language models to access as much information as possible, necessitating a method that can effectively store and retrieve information.

An essential direction for improving long-context processing involves information compression, encapsulating prior key-value (KV) information within a few specialized tokens. Previous efforts, such as \cite{mu_learning_2023}, have achieved this goal with relative efficacy. However, a notable limitation of these methods is their lossy nature of compression, which inevitably leads to the loss of vital information during the process.

We propose a novel approach, the Lossless Compressed Memory Attention (LoMA), which divides sequence into multiple chunks of equal length, each chunk structured to include a reading zone, a memory zone and a repetition zone. The latter two zones incorporate newly introduced special tokens: `\textless m\textgreater' and `\textless r\textgreater'. We also designed a unique attention matrix mask: the reading zone employs a conventional autoregressive lower triangular mask; in order to facilitate better internal information transmission and communication, the memory zone employs a bidirectional attention mechanism and they can attend to reading zone; tokens in the repetition zone can only observe the memory zone directly preceding it, as well as the token itself. With this masking strategy, the `\textless r\textgreater' token in the repetition zone needs to faithfully reproduce the text content of the reading zone, while only being able to attend to the <m> tokens in the memory zone. This implies that the `\textless m\textgreater' tokens quickly learn to compress the entire content of the reading zone into their own KV.

We have also mathematically demonstrated that the loss function generated in the repetition zone can indirectly supervise the training of the model in the memory zone, obviating the need for constructing labels and computing loss for the tokens in the memory zone.

Through the generative algorithm of LoMA, transformer models acquire the ability to compress memory losslessly within the memory zone, substantially extending the length of the long-context they are capable of handling and significantly reducing computational and memory costs. Our experiments show that the Llama-2-7B model\cite{touvron_llama_2023}, when fine-tuned with the LoMA training method, is capable of high-ratio lossless memory compression of its own KV cache. Importantly, our approach does not modify the model's architecture or rely on additional auxiliary models.

Chapter 2 reviews several studies related to our methodology, Chapter 3 provides an in-depth explanation of the LoMA generation algorithm, Chapter 4 describes the training precedure for endowing the transformer model with memory compression capabilities, Chapter 5 discusses our experimental results, and Chapter 6 concludes with a summary of our work.

\section{Related Works}
\label{related}

\subsection{Sparse Attention}

In recent times, the computational burden of long contexts has been effectively alleviated with the introduction of various sparsified attention mechanisms. \cite{zaheer_big_2021} integrating random attention, windowed attention, and global attention achieved commendable results. \cite{zhao_explicit_2019}, \cite{gupta_memory-efficient_2021} posits that the plethora of irrelevant information within the attention mechanism can be distracting for the model, and thus zeroes out the less significant positions within the attention matrix to focus the model's attention. Subsequently, \cite{zhang_h_2o_2023} proposed a method to filter tokens of importance by summing up attention scores. Going a step further, \cite{ribar_sparq_2023} estimated attention scores in the embedding dimension using the top-r values to then select the top-k largest KV pairs. The recently prominent Mistral architecture\cite{jiang_mistral_2023}, employs windowed attention akin to the receptive fields of CNNs\cite{oshea_introduction_2015}, theoretically enabling the effortless handling of text sequences up to the length of 32 $\times$ 4096. However, none of these works can achieve lossless compression of context. More or less, some important information will be lost.

\subsection{Explicit Memory}

Explicit memory is the conscious, intentional recollection of factual information, previous experiences, and concepts. Some method for Explicit memory compression are proposed by \cite{lanchantin_learning_2023}, \cite{jiang_llmlingua_2023}. Those approach involves the generation of a summary of preceding text, which is then inserted into the generated text, allowing subsequent text generation to utilize this summary to produce more coherent text. The downsides of this method include: 1) the generated summary occupies a significant portion of the text length, resulting in shorter generated text; 2) the process of generating a summary is also autoregressive, leading to a substantial increase in generation time; 3) the generated summary may omit some critical information, compromising the accuracy of the resulting text; and 4) a considerable amount of annotated data is required to fine-tune the model, which is costly.

In \cite{mu_learning_2023}, a novel compression method was introduced. This method involves inserting a `gist token' between the prompt and response and employing a specially designed mask to ensure that the response chunk can only extract information from the gist token. During generation, the prompt is compressed into a gist token and then the original prompt is discarded to save resources. This approach effectively reduces memory usage. However, it's important to note that this method is not lossless and results in a significant loss of information. In contrast, our method achieves lossless compression of information into a `\textless m\textgreater' token, ensuring that no information is lost.

\begin{figure}[]
    \centering
    \includegraphics[width=\columnwidth]{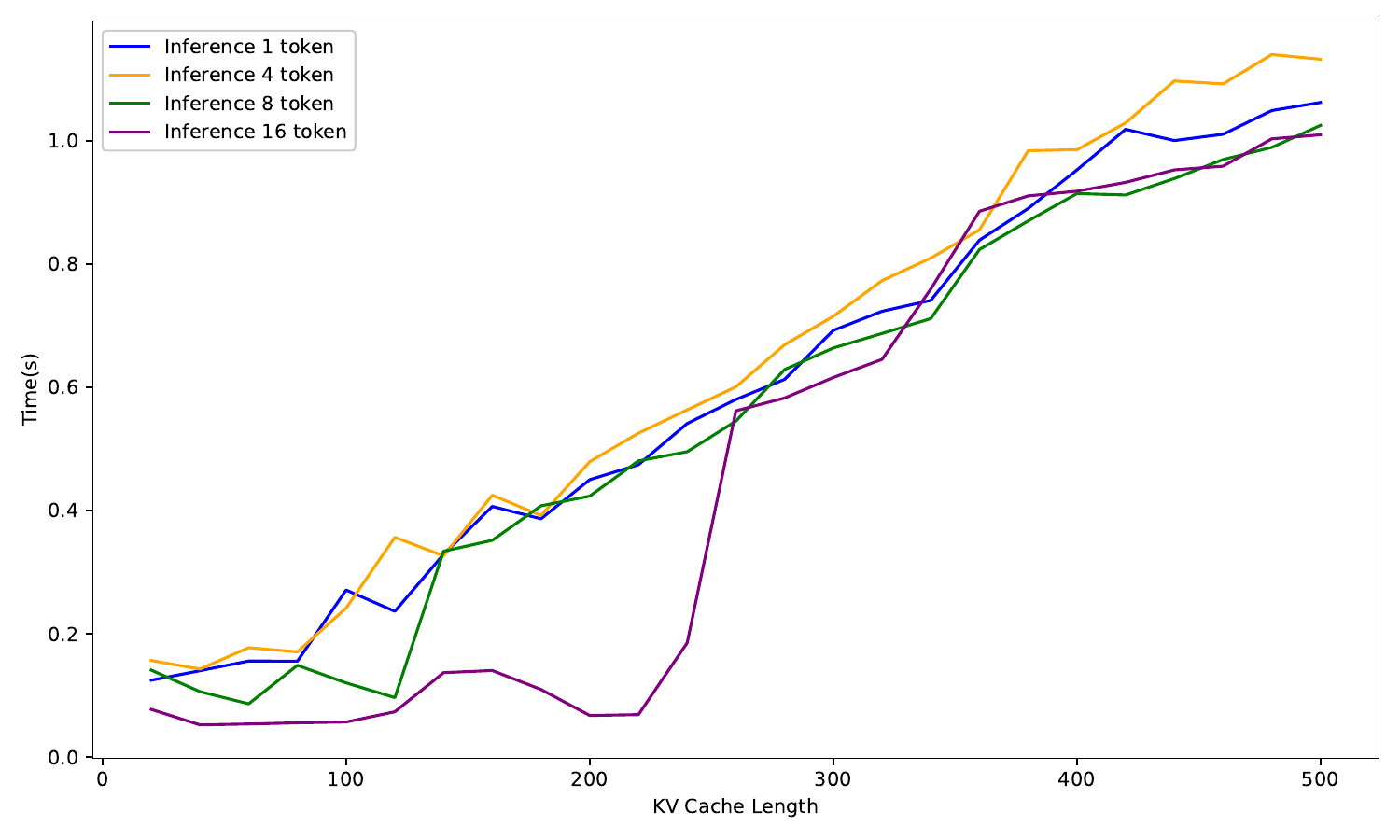}
    \caption{This figure delineates the relationship between single inference latency and KV cache length across various input token sequence lengths. The findings indicate that the latency of a single inference grows linearly with the length of the KV cache, yet the augmentation of input token sequence length does not substantially affect the computation time. Notably, when the input sequence consists of 16 tokens, an increase in KV cache length from 0 to 240 does not incur additional inference time, which might be attributable to the computational capacity characteristics of the hardware.}
    \label{fig_8}
\end{figure}

\section{Method}

The LoMA framework introduces an enhanced autoregressive generation algorithm that leverages a transformer model trained to compress the KV cache losslessly. We first detail this algorithm and then describe the training methodology necessary to imbue the model with this advanced capability.

\subsection{LoMA Generation}

Within the architecture of a transformer, the KV (key-value) cache stores information from the preceding context and integrates it into the computation of attention. As the generated sequence lengthens, the memory occupied by the KV cache increases proportionally, leading to greater computational costs. Our proposed method, Lossless Compressed Memory Attention (LoMA), introduces an efficient computation step within the generation process to execute high-ratio lossless compression on the KV-cache. This significantly curtails storage and computational resource usage.

LoMA functions with a defined compression ratio $c$ and a target compressed length $t$. Within the enhanced autoregressive generation framework, once the model accumulates a KV cache spanning $tc$ tokens, LoMA model compresses it to a fixed length $t$, as illustrated in Fig\ref{fig_6} (b). This compression is achieved through the following steps:

\begin{enumerate}[itemsep=-1pt,topsep=0pt,parsep=0pt]
\item The model employs a standard autoregressive generation process to produce a sequence of $tc$ tokens, yielding a KV cache of corresponding length. This particular subset of tokens forms the \textit{reading zone}, which is denoted by $\text{KV}_{\text{Read}}$.
\item A single inference pass is conducted on $t$ `\textless m\textgreater' tokens with $\text{KV}_{\text{Read}}$, which yields a condensed KV cache of length $t$. This subsequence is designated as the \textit{memory zone}.
\item The reading zone's KV cache is discarded, and following autoregressive generation proceeds utilizing the compressed KV cache from the memory zone.
\end{enumerate}

A comprehensive code listing detailing the aforementioned steps is presented in Appendix \ref{Loma_generator}.

\subsection{Performance analysis}

In this analysis, we evaluated the extent to which LoMA reduces the computational and storage resource requirements. Without loss of generality, we compared the standard autoregressive generation algorithm with the LoMA generation in the absence of prompts. Let $T_\text{infer}(l, k)$ denote the time it takes for the model to complete one inference on a token sequence of length $l$ with a key-value (KV) cache of length $k$. Assuming the total generation spans $m$ chunks, each consisting of $tc$ tokens, the generation time for a traditional transformer is given by:

$$
\sum_{k=0}^{mtc-1}{T_\text{infer}(1, k)}
$$

Under a preset compression ratio $c$ and memory length $t$, LoMA performs one inference every $tc$ tokens with $t$ `\textless m\textgreater' tokens, resulting in a total generation time of:

$$
\sum_{y=0}^m{\sum_{k=yt}^{yt+tc-1}{T_\text{infer}(1, k)}} + mT_\text{infer}(tc, t)
$$

Typically, $T_\text{infer}(l, k)$ is much less than $lT_\text{infer}(1, k)$. Our tests conducted on an A100 GPU demonstrate this point, see Fig.\ref{fig_8}.

Consequently, even though the additional term $mT_\text{infer}(tc, t)$ of the LoMA generation process slightly increases the computation, the significant compression of the KV cache results in a notable reduction in both generation time and memory usage, as illustrated in the table \ref{tab_3}.

\begin{figure}[]
    \centering
    \includegraphics[width=\columnwidth]{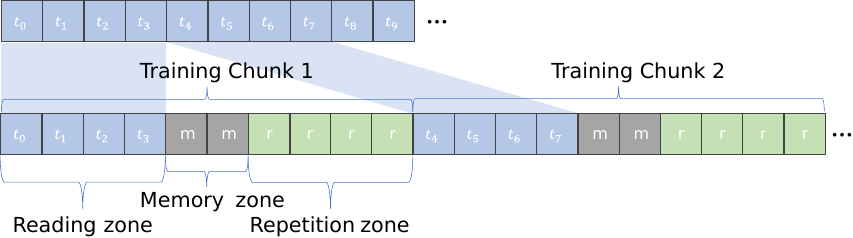}
    \caption{The top row represents the original training samples, while the bottom row shows the processed training samples used for training or fine-tuning the LoMA model. In the original training samples, we insert $t$ `\textless m\textgreater' tokens and $tc$ `\textless r\textgreater' tokens after every $tc$ tokens.}
    \label{fig_7}
\end{figure}

\section{Training}
\label{Training}

To equip the transformer model with the aforementioned memory compression capability, pre-training or fine-tuning procedures are essential. We have devised a training procedure that includes structured reorganization of input samples, a novel loss terms, a unique design attention mask, and a specialized pattern of PositionIDs.

\subsection{Input Samples}

In the training procedure of LoMA, the original sequence of tokens is segmented into multiple subsequences each of length $tc$. To each subsequence, $t$ `\textless m\textgreater' tokens followed by $tc$ `\textless r\textgreater' tokens are appended, forming a \textit{training chunk}. All training chunks are concatenated to form a new structured sequence as a training sample. see Fig \ref{fig_7}.

\subsection{Loss}

To train a Transformer model using the structured input sequence mentioned above, it is necessary to extend the vanilla loss $\mathcal{L}_{\text{LM}}$ with an additional term that endows the model with the capability to compress memory. Since the output of model on the memory zone is not of concern, with the KV-cache in this zone being utilized to store compressed information, there is no need to design labels or a loss function for the memory zone. Indirectly, the memory zone is supervised through the loss applied to the repetition zone. Consequently, the training loss for each chunk is calculated as the sum of these two components, and the total loss across all chunks is determined by:

\begin{equation}
    \mathcal{L} = \sum_{y=1}^{m}{\left(\mathcal{L}_{\text{Read}}^y + \mathcal{L}_{\text{Rep}}^y\right)}
    \label{equ_11}
\end{equation}

where $\mathcal{L}_{\text{Read}}^y$ is the loss generated by the reading zone of the $y$-th training chunk while $\mathcal{L}_{\text{Rep}}^y$ corresponds to the loss produced by the repetition zone.

\begin{figure}[]
    \centering
    \includegraphics[width=\columnwidth]{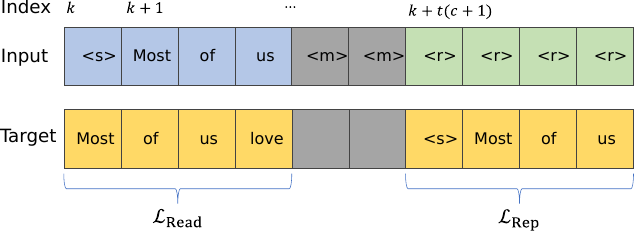}
    \caption{This figure describes the correspondence between inputs and labels. In reading zone, the input and target exhibit a standard autoregressive relationship. No labels are set in the memory zone, while the labels in the repetition zone consist of content from the reading zone. We demonstrated in Section.\ref{Gradient} that by backpropagating gradients through the repetition zone, a supervisory signal can be provided to the memory zone. This allows the `\textless m\textgreater' token to learn to compress the content of the reading zone into its own KV.}
    \label{fig_9}
\end{figure}

Let the token subsequence from the reading zone be denoted as $\text{READ}_y=\left\{x_k,x_{k+1},...,x_{k+tc}\right\}$, we have:

\begin{equation}
\mathcal{L}_{\text{Read}}^y=\sum_{i=k}^{k+tc}{\text{CE}\left(M(x_i),x_{i+1}\right)}
\label{equ_read_loss}
\end{equation}

is same with the standard trainning loss and

\begin{equation}
\mathcal{L}^y_{\text{REP}}=\sum_{i=k}^{k+tc}{\text{CE}\left(M(\text{`\textless r\textgreater'}_{i+t(c+1)}), x_i\right)}
\label{equ_rep_loss}
\end{equation}

where $\text{CE}(\text{logits}, \text{label})$ refers to the standard cross-entropy loss function, and the $M(x)$ is the set of logits produced by the model $M$ for the token $x$. The term $\text{`\textless r\textgreater'}_{k+t(c+1)}$ refers to the `\textless r\textgreater' token at position $k+t(c+1)$, indicating that the model's prediction for each `\textless r\textgreater' token should be identical with the corresponding token in the reading zone. Refer to Fig.\ref{fig_9} for visual clarification.

\begin{figure}[]
    \centering
    \includegraphics[width=0.90\columnwidth]{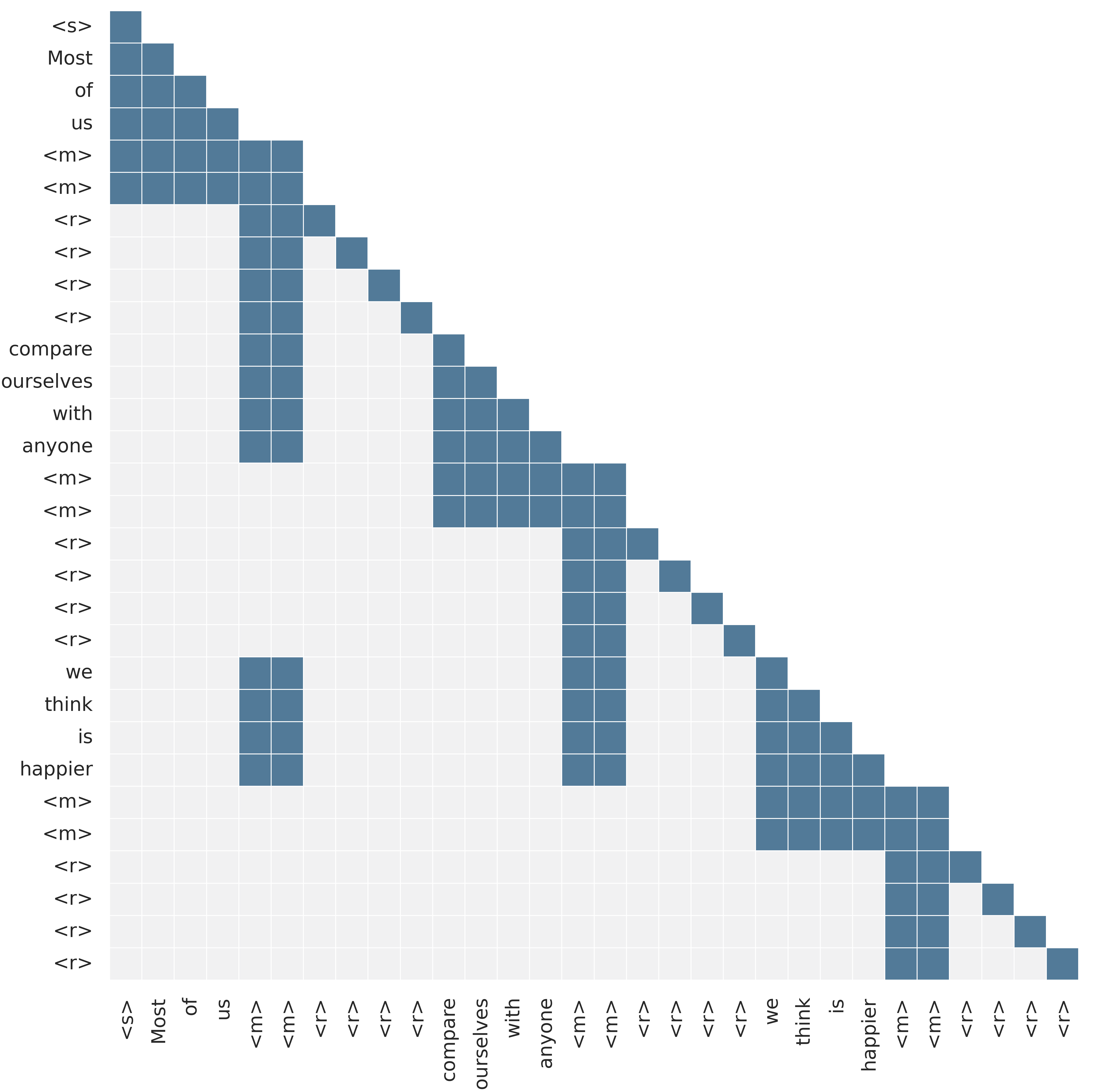}
    \caption{The figure presents an attention mask for an input sequence comprising 12 tokens, which includes the initial token `\textless s\textgreater'. In this configuration, with $t=2$ and $c=2$, the reading and repetition zones each span 4 tokens, and the recall zone encompasses 2 tokens. Accordingly, the sequence is segmented into three training chunks. Each chunk is prefixed with `\textless m\textgreater' and suffixed with `\textless r\textgreater' tokens, yielding a total chunk length of 10 tokens (4+2+4). This results in an attention mask with a dimension of $30\times30$. Within this matrix, grey squares indicate a value of $0$, which blocks attention, and blue squares represent a value of $1$, allowing attention to flow.}
    \label{fig_1}
\end{figure}

\subsection{Mask}

Given that the input token sequence is restructured into training chunks, we cannot employ the standard lower triangular masking matrix directly in training. Instead, we must redesign the masking matrix to meet the following criteria:

\begin{enumerate}[itemsep=-1pt,topsep=0pt,parsep=0pt]
    \item To enable the model to extract information from the memory zones for generation, the mask for the reading zone is designed to attend to all preceding memory zones, while maintaining unidirectional attention within the reading zone of the current training chunk.
    \item The attention mask designed for the memory zone requires bidirectional attention. This facilitates the interaction of the KV cache, with a span of $t$ in the memory zone, effectively condensing the storage requirements for the KV cache of length $tc$ in the reading zone. It is also critical to prevent the memory zone from attending to any tokens beyond the adjacent reading zone.
    \item In the repetition zone, the mask is designed to ensure that each token is restricted to attending only to the preceding memory zone. This constraint is vital because the repetition zone must reconstruct content based solely on the memory zone's information, without depending on the previously 'recalled' tokens to 'generate' subsequent tokens.
\end{enumerate}

For a training chunk composed of a reading zone of length $tc$, a memory zone of length $t$, and a repetition zone of length $tc$, the attention mask is defined as:

\begin{equation}
    \textbf{M}_{s \times s}=\left[\begin{matrix}
        \textbf{L}_{tc\times tc}&\textbf{0}_{tc\times t}&\textbf{0}_{tc\times tc} \\
        \textbf{1}_{t\times tc}&\textbf{1}_{t\times t}&\textbf{0}_{t\times tc} \\
        \textbf{0}_{tc\times tc}&\textbf{1}_{tc\times t}&\textbf{I}_{tc\times tc}
    \end{matrix}\right]\label{equ_1}
\end{equation}

Where $s=t(2c+1)$. During training, all training chunks are concatenated to form a single training sample, and the attention masks for each chunk are likewise concatenated in the following manner to form the attention mask for the entire sample.

\begin{equation}
    \textbf{M}=\left[\begin{matrix}
        \textbf{M}_{s\times s}&\textbf{0}_{s\times s}&\textbf{0}_{s\times s}&\ldots&\textbf{0}_{s\times s} \\
        \textbf{S}_{s\times s}&\textbf{M}_{s\times s}&\textbf{0}_{s\times s}&\ldots&\textbf{0}_{s\times s} \\
        \textbf{S}_{s\times s}&\textbf{S}_{s\times s}&\textbf{M}_{s\times s}&\ldots&\textbf{0}_{s\times s} \\
        \vdots&\vdots&\vdots&\ddots&\vdots \\
        \textbf{S}_{s\times s}&\textbf{S}_{s\times s}&\textbf{S}_{s\times s}&\ldots&\textbf{M}_{s\times s}
    \end{matrix}\right]\label{equ_2}
\end{equation}

Where

\begin{equation}
    \textbf{S}_{s \times s}=\left[\begin{matrix}
        \textbf{0}_{tc\times tc}&\textbf{1}_{tc\times t}&\textbf{0}_{tc\times tc} \\
        \textbf{0}_{t\times tc}&\textbf{0}_{t\times t}&\textbf{0}_{t\times tc} \\
        \textbf{0}_{tc\times tc}&\textbf{0}_{tc\times t}&\textbf{0}_{tc\times tc}
    \end{matrix}\right]\label{equ_3}
\end{equation}

In this design, the 1s in \textbf{S} allow the reading zone to attend to the KV caches of all preceding memory zones. Fig.\ref{fig_1} illustrates the attention mask $M$ for a complete sample.

\begin{table*}[]
\caption{Resource Savings in Computational and Memory Footprint at Various $t$ and $c$ Parameter Configurations}
\begin{tabular}{@{}ccccccc@{}}
\toprule
 & \multicolumn{2}{c}{t=4}    & \multicolumn{2}{c}{t=8}    & \multicolumn{2}{c}{t=16}   \\ \midrule
  ratio    & comp. overhead & memory reduced & comp. overhead & memory reduced & comp. overhead & memory reduced \\ \midrule
2     & $-43.82\%$          & $-200\%$         & $-45.35\%$         & $-200\%$         & $-47.79\%$         & $-200\%$         \\
4     & $-65.87\%$         & $-400\%$         & $-67.12\%$         & $-400\%$         & $-66.95\%$         & $-400\%$         \\
8     & $-75.19\%$         & $-800\%$         & $-74.24\%$         & $-800\%$         & $-71.41\%$        & $-800\%$         \\ \bottomrule
\end{tabular}
\label{tab_3}
\end{table*}

\subsection{Gradient}
\label{Gradient}

In this section, we will demonstrate that LoMA can still learn the ability to compress memory during training, even without  supervising the output of `\textless m\textgreater' tokens.

Suppose the input sequence is organized into a reading zone $r$ of length $tc$, a memory zone $m$ of length $t$, and a repetition zone $p$ of the same length as the reading zone, making the total length $s=t(2c+1)$. Therefore, Q, K, and V are each composed of three vectors concatenated together as Equ.\ref{equ_4}.

\begin{equation}
    Q=\begin{pmatrix}q_r\\q_m\\q_p\end{pmatrix},K=\begin{pmatrix}k_r\\k_m\\k_p\end{pmatrix},V=\begin{pmatrix}v_r\\v_m\\v_p\end{pmatrix}
    \label{equ_4}
\end{equation}

By substituting the above expressions into the formula for Attention, we get the following calculation as Equ.\ref{equ_5}.

\begin{equation}
    \hat{\textbf{A}}=\begin{bmatrix}(q_rk_r^T)_{tc \times tc}&(q_rk_m^T)_{tc \times t}&(q_rk_p^T)_{tc \times tc}\\(q_mk_r^T)_{t \times tc}&(q_mk_m^T)_{t \times t}&(q_mk_p^T)_{t \times tc}\\(q_pk_r^T)_{tc \times tc}&(q_pk_m^T)_{tc \times t}&(q_pk_p^T)_{tc \times tc}\end{bmatrix}
    \label{equ_5}
\end{equation}

Dot product $\hat{\textbf{A}}$ with the previously designed mask (Equ.\ref{equ_1}) as Equ.\ref{equ_6}.
\begin{equation}
\textbf{A}=\hat{\textbf{A}}\cdot \textbf{M}_{s\times s}=\begin{bmatrix}q_rk_r^T\odot \textbf{L}&\textbf{0}&\textbf{0}\\q_mk_r^T&q_mk_m^T&\textbf{0}\\\textbf{0}&q_pk_m^T&q_pk_p^T\odot \textbf{I}\end{bmatrix}
    \label{equ_6}
\end{equation}

Here, $\odot$ denotes element-wise multiplication, $\textbf{L}$ is a lower triangular mask matrix. Without loss of generality, if we ignore the scale factor$\sqrt{d_k}$, then the output of the Attention Block can be expanded as Equ.\ref{equ_7}.

\begin{equation}
\begin{aligned}\textbf{O}&=\text{Softmax}(\textbf{A})V \\&=\begin{pmatrix}E_1\exp(\textbf{A}_1)V\\E_2\exp(\textbf{A}_2)V\\...\\E_s\exp(\textbf{A}_s)V\end{pmatrix}\end{aligned}
    \label{equ_7}
\end{equation}

Where  $\textbf{A}_i$ represents the $i$-th row vector of $\textbf{A}$, and $E_i=1/\sum{\exp(\textbf{A}_i)}$ . Since we are currently only studying the propagation of gradients and not concerned with the absolute magnitude of the gradients, we can ignore coefficients(i.e., $E_i$).  Thus, the first  $tc$  rows of $\textbf{O}$, which represent the output of the reading zone, can be expressed as Equ.\ref{equ_8}.

\begin{equation}
\textbf{O}_r=\exp(q_rk_r^T\odot \textbf{L})v_r
    \label{equ_8}
\end{equation}

The rows from $s-tc$ to $s$ of $\textbf{O}$, which represent the output of the repetition zone, can be expressed as Equ.\ref{equ_9}

\begin{equation}
\textbf{O}_p=\exp(q_pk_m^T)v_m + \exp(q_pk_p^T\odot \textbf{I})v_p
    \label{equ_9}
\end{equation}

Since we do not supervise the output of `\textless m\textgreater' tokens, the loss consists only of two parts: $\mathcal{L}_\text{LM}$ for the reading zone and $\mathcal{L}_\text{repeat}$. Therefore, the gradient of the loss with respect to $x$ is as follows:
\begin{equation}
\frac{\partial\mathcal{L}}{\partial x}=\frac{\partial\mathcal{L}_\text{Read}}{\partial\textbf{O}_r}\cdot\frac{\partial\textbf{O}_r}{\partial x}+\frac{\partial\mathcal{L}_\text{Rep}}{\partial\textbf{O}_p}\cdot\frac{\partial\textbf{O}_p}{\partial x}
    \label{equ_10}
\end{equation}

It's important to note that: in Equ.\ref{equ_10},  $\textbf{O}_p$ is a function of $k_m$ and $v_m$, indicating that the model's memory capability can receive supervisory signals from the loss in the repetition zone.

\subsection{Position IDs}

Due to the intrinsic structure of the training samples in the LoMA model, which significantly differ from the original plain-form samples, it is necessary to design a corresponding set of position IDs. Assuming the original input sequence is segmented into chunks of length $t$, the starting position ID for the $i$-th chunk is $it$. After organizing this chunk into a training chunk of length $t(2c+1)$, the position IDs for the memory zone are designed as:

\begin{equation}
    it+c-1,it+2c-1,...,it+tc-1
    \label{equ_14}
\end{equation}

The purpose of this design is twofold: 1) to preserve the original position ids of the reading zone, thereby maintaining consistent position encodings and ensuring coherence in the KV cache, and 2) to make the position ids of the memory zone contiguous with those of the reading zone in the subsequent chunk, facilitating information extraction.

\begin{figure*}[htbp]
    \centering
    \includegraphics[width=\textwidth]{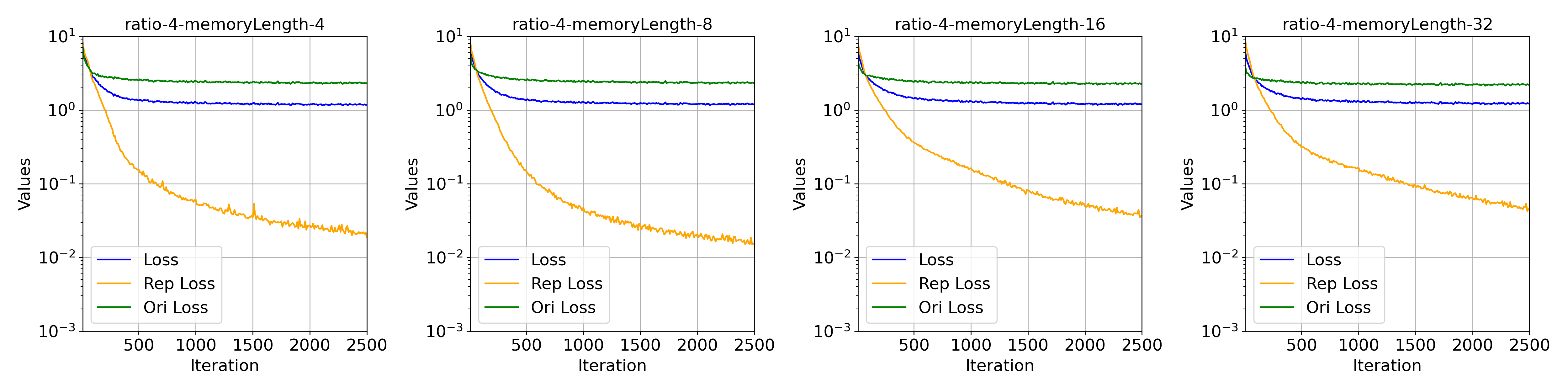}
    \caption{The comparison of different compression ratios on same length of memory zone. The orange lines represent $\mathcal{L}_{\text{Rep}}$ from Equ.\ref{equ_11}, the green lines represent $\mathcal{L}_{\text{Read}}$, and the blue lines represent the total loss $\mathcal{L}$.}
    \label{fig_2}
\end{figure*}

\begin{figure*}[htbp]
    \centering
    \includegraphics[width=\textwidth]{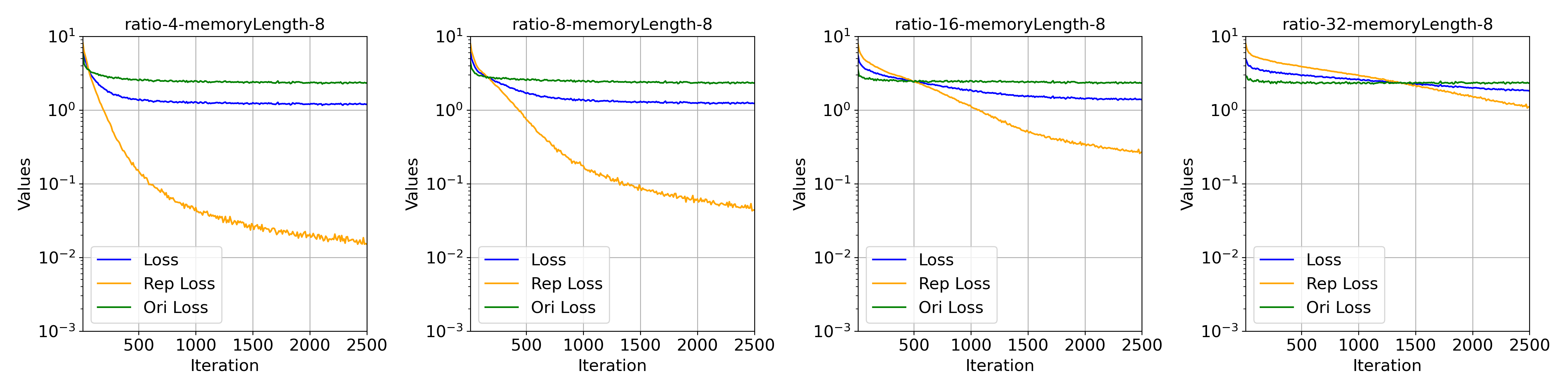}
    \caption{The comparison of different lengths of memory zone on same compression ratio.}
    \label{fixed_memlen}
\end{figure*}


\section{Experiments}

\subsection{Settings}

\textbf{Model.} We conduct experiments based on the pretrained Llama2 7B Chat models. The training framework is Megatron-LM\cite{shoeybi_megatron-lm_2020}, using a machine equipped with 8 A100 GPUs. The Global batch-size are set to 32. The learning rate adopts a cosine decay pattern, varying between $[5e-5, 5e-6]$, and includes a warmup of 100 iterations.

\textbf{Special Tokens.} The tokenizer of Llama2 has a vocabulary length of $32,000$. We added two special tokens with IDs `\textless m\textgreater: 32000' and `\textless r\textgreater: 32001'. Accordingly, modifications are needed in the model’s embedding layer: two vectors of $4096$ dimensions are added, changing the weight shape from $32000\times 4096$ to $32002\times 4096$. These two new vectors are initialized with a normal distribution, the mean and variance of which are derived from the statistics of the original weights.

\textbf{Framework.} Since LoMA requires a custom attention mask and position id, we are unable to use Flash-Attention to accelerate the training and need to make certain modifications to Megatron\cite{shoeybi_megatron-lm_2020}.

\textbf{Data.} A small portion is extracted from the C4 dataset\cite{raffel_exploring_2020} or the GSM8K dataset\cite{cobbe_training_2021} is used in a loop to train the model. The method of data preprocessing is as mentioned in the previous section.\ref{Training}. Assuming $\hat{s}$ is the preset training sequence length, we impose the following requirement:

$$
\nonumber
\hat{s} \text{ mod } \left(2tc + t \right) = 0
$$
We divide the training space into $\hat{s} / \left(2tc + t \right)=\bar{n}$ chunks, adding memory and repetition zones to each chunk. Assuming the sequence length input into the model is $s$, then it is required that:
$$
2s + \left\lfloor \frac{s}{c}\right\rfloor \leq \hat{s}
$$

where, $c$ is the compress ratio. If $s+\bar{n}t(c+1)<\hat{s}$, it needs to be padded to the length of $\hat{s}$.

\begin{table*}[]
    \centering
    \caption{Repetition Accuracy}
    \begin{tabular}{@{}ccccc@{}}
        \toprule
        \textbf{Loma Hyperparameters} & $c=4,t=8$ & $c=4,t=16$ & $c=8,t=8$ & $c=8,t=16$ \\ \midrule
         Zone accuracy & \textbf{71.56\%} & 54.90\% & 35.30\% & 40.19\% \\
         Token accuracy & \textbf{99.84\%} & 99.68\% & 99.38\% & 99.40\% \\ \bottomrule
    \end{tabular}
    \label{tab_1}
\end{table*}

\subsection{Main Results}

Firstly, we tested different compression ratios while keeping the length of memory zone constant at $t=8$. The training data for this experiment was the C4, and the LLM used was Llama-2-7b. During training, it was observed that $\mathcal{L}_{\text{Rep}}$ decreased rapidly and eventually approached zero when $c \leq 8$. The model was able to nearly perfectly recapitulate the content of the reading zone relying solely on the significantly compressed memory zone's KV cache, demonstrating the model's capability for lossless compression of the KV cache. See Fig. \ref{fig_2}

Additionally, under the same experimental conditions and a constant compression ratio of $c=4$, we compared the effects of varying the $t$ values. The experimental findings indicate that changes in the $t$ value (ranging from 4 to 32) have a minimal impact on the model's memory compression capabilities, as illustrated in the Fig. \ref{fixed_memlen}.

Subsequently, across multiple different parameter settings, we trained on the C4 dataset for 5000 iterations and then conducted inference testing on GSM8K. We evaluated the generalization of the model's memory compression ability by calculating the accuracy within the repetition zone. In this experiment, we defined two distinct accuracy metrics: zone accuracy and token accuracy. Zone accuracy denotes the percentage of zones that were recapitulated entirely correctly, while token accuracy represents the percentage of tokens that were accurately recapitulated. See Tab.\ref{tab_1}.

\section{Conclusion}

We propose the Lossless Compressed Memory Attention (LoMA), aimed at losslessly compressing information to reduce computational consumption in long text contexts. The advantages of this approach are: 1) It does not alter the model structure, allowing for an expansion of the model's contextual length to $c$ times its original size for most models; 2) It does not require additional annotated data and can be fine-tuned directly on pre-trained models; 3) It allows for segmental compression, and each compression only adds one inference process, avoiding a significant increase in generation time. We fine-tuned the LLaMA 7B model with LoMA on the C4 and GSM8K datasets, achieving convergence within 2000 iterations. Moreover, we found that information compression has good generalizability; models trained on C4 can be seamlessly generalized to the GSM8K dataset. We suggest adopting LoMA in pretraining to address the increasingly important scenarios of long texts in the future.

\bibliography{main}
\bibliographystyle{icml2024}
\onecolumn

\newpage
\appendix
\section{Loma generator}
\label{Loma_generator}

\begin{lstlisting}
class LoMAGenerator:
    def __init__(self, model, max_len, position_type, compress_ratio, mem_len, mem_token_id) -> None:
        self.model = model
        self.model_dtype = model.parameters().__next__().dtype
        self.max_len = max_len
        self.mem_len = mem_len
        self.compress_ratio = compress_ratio
        self.position_type = position_type
        self.read_len = compress_ratio * mem_len
        self.mem_token_id = mem_token_id
        self.reset()
        self.position_ids = None

    def reset(self):
        self.cursor = 0
        self.kv_cache = None
        self.generated = []
        self.compressed_chunks = 0
        self.input_buffer = []
        self.position_ids = None

    def __call__(self, input_ids, eos_token_id):
        generated = []
        generated.append(self.add_token_ids(input_ids))
        while len(generated) < self.max_len and generated[-1] != eos_token_id:
            generated.append(self.add_token_ids([generated[-1]]))
        return generated

    def add_token_ids(self, token_ids):
        assert len(token_ids) > 0
        self.input_buffer.extend(token_ids)
        while self.mem_len > 0:
            # WHILE loop iterates inference and compression
            kv_cache_len = self.kv_cache[0][0].shape[2] if self.kv_cache else 0
            uncomp_cache_len = kv_cache_len - self.compressed_chunks * self.mem_len
            assert uncomp_cache_len < self.read_len
            proc_len = self.read_len - uncomp_cache_len
            if len(self.input_buffer) < proc_len:
                break
            input_ids = self.input_buffer[:proc_len]
            self.input_buffer = self.input_buffer[proc_len:]
            last_predict = self.inference_input(input_ids)
            self.compress_last_chunk()
        if self.input_buffer:
            last_predict = self.inference_input(self.input_buffer)
            self.input_buffer = []
        return last_predict

    def inference_input(self, input_ids):
        input_ids = torch.LongTensor(input_ids)
        position_ids = torch.LongTensor(range(self.cursor, self.cursor + len(input_ids)))
        kv_cache_len = 0 if self.kv_cache is None else self.kv_cache[0][0].shape[2]
        mask_rows = input_ids.shape[0]
        mask_cols = mask_rows + kv_cache_len
        attn_mask = torch.ones((1, 1, mask_rows, mask_cols)).cuda().to(self.model_dtype)
        attn_mask = torch.tril(attn_mask, diagonal=kv_cache_len)
        with torch.no_grad():
            output = self.model(
                input_ids = input_ids.cuda().unsqueeze(0),
                past_key_values = self.kv_cache,
                position_ids = position_ids.cuda().unsqueeze(0),
                attention_mask = attn_mask,
                return_dict = True,
                )
        self.kv_cache = output['past_key_values']
        self.cursor += input_ids.shape[0]
        return output['logits'][:, -1, :].argmax(dim=-1).item()

    def compress_last_chunk(self):
        """
        Compress all the read KV cache into the KV of <m> tokens
        """
        assert self.kv_cache is not None
        mem_cursor = self.compressed_chunks * self.mem_len
        cache_len = self.kv_cache[0][0].shape[2]
        assert mem_cursor == cache_len - self.read_len

        input_ids = torch.LongTensor([self.mem_token_id] * self.mem_len)
        if self.position_type == 'intermittent':
            position_ids = list(range(
                self.cursor - self.read_len + self.compress_ratio - 1,
                self.cursor,
                self.compress_ratio
                ))
        else:
            position_ids = list(range(mem_cursor, mem_cursor + self.mem_len))
        position_ids = torch.LongTensor(position_ids)

        read_kv_cache = []
        for i in range(len(self.kv_cache)):
            cache_k, cache_v = self.kv_cache[i]
            read_kv_cache.append((cache_k[:, :, -self.read_len:],
                                  cache_v[:, :, -self.read_len:]))
        with torch.no_grad():
            mem_attn_mask = torch.ones(1, 1, self.mem_len, self.mem_len + self.read_len).to(self.model_dtype)
            output = self.model(
                input_ids = input_ids.cuda().unsqueeze(0),
                past_key_values = read_kv_cache,
                position_ids = position_ids.cuda().unsqueeze(0),
                attention_mask = mem_attn_mask.cuda(),
                return_dict = True,
                )

        read_kv_cache = output['past_key_values']
        if isinstance(self.kv_cache, tuple):
            self.kv_cache = list(self.kv_cache)
        for i in range(len(read_kv_cache)):
            old_cache_k, old_cache_v = self.kv_cache[i]
            mem_cache_k, mem_cache_v = read_kv_cache[i]
            self.kv_cache[i] = (
                torch.cat([old_cache_k[:, :, :mem_cursor], mem_cache_k[:, :, -self.mem_len:]], dim=2),
                torch.cat([old_cache_v[:, :, :mem_cursor], mem_cache_v[:, :, -self.mem_len:]], dim=2)
            )
        self.compressed_chunks += 1
\end{lstlisting}

\end{document}